\documentclass[10pt,twocolumn,letterpaper]{article}

\usepackage{wacv}              

\usepackage{graphicx}
\usepackage{amsmath}
\usepackage{amssymb}
\usepackage{booktabs}
\usepackage{times}
\usepackage{caption}
\usepackage{subcaption}
\usepackage{multirow}
\usepackage{bm}
\usepackage{balance}
\usepackage{enumitem}
\usepackage[accsupp]{axessibility}
\usepackage[linesnumbered,ruled,lined]{algorithm2e}
\graphicspath{ {./images/} }

\usepackage[pagebackref,breaklinks,colorlinks]{hyperref}

\usepackage[capitalize]{cleveref}
\crefname{section}{Sec.}{Secs.}
\Crefname{section}{Section}{Sections}
\Crefname{table}{Table}{Tables}
\crefname{table}{Tab.}{Tabs.}

\def\wacvPaperID{*****} 
\def\confName{WACV}
\def\confYear{2024}

\begin{document}

\title{Alpha-wolves and Alpha-mammals: Exploring Dictionary Attacks on Iris Recognition Systems}

\author{Sudipta Banerjee\textsuperscript{1}, Anubhav Jain\textsuperscript{1}, Zehua Jiang\textsuperscript{1}, Nasir Memon\textsuperscript{1}, Julian Togelius\textsuperscript{1}, Arun Ross\textsuperscript{2}\\
\textsuperscript{1}New York University\\
{\tt\small \{sb9084, aj3281, zj2086, memon, julian.togelius\}@nyu.edu}
\and
\textsuperscript{2}Michigan State University\\
{\tt\small rossarun@msu.edu}
}

\maketitle

\begin{abstract}
A dictionary attack in a biometric system entails the use of a small number of strategically generated images or templates to successfully match with a large number of identities, thereby compromising security. We focus on dictionary attacks at the template level, specifically the IrisCodes used in iris recognition systems. We present an hitherto unknown vulnerability wherein we mix IrisCodes using simple bitwise operators to generate alpha-mixtures \textemdash alpha-wolves (combining a set of ``wolf" samples) and alpha-mammals (combining a set of users selected via search optimization) that increase false matches. We evaluate this vulnerability using the IITD, CASIA-IrisV4-Thousand and Synthetic datasets, and observe that an alpha-wolf (from two wolves) can match upto 71 identities @FMR=0.001\%, while an alpha-mammal (from two identities) can match upto 133 other identities @FMR=0.01\% on the IITD dataset. 
\end{abstract}

\section{Introduction}
In a dictionary attack, a small number of biometric samples or templates are strategically generated such that they successfully match with a large number of identities. Dictionary attacks on biometric recognition systems were first described in the context of fingerprints~\cite{MasterPrint}, where the authors demonstrated the vulnerability of  {\em small-sized} sensors that enroll {\em multiple} low-resolution fingerprint samples. The authors synthesized fingerprint ``templates" using a brute force approach that could match a large proportion of identities in an unseen population. They further devised the latent variable evolution method to generate \textit{Deep MasterPrints}~\cite{DeepMP}. Following the success of masterprints, the feasibility of dictionary attacks in other biometric modalities were explored. \textit{MasterFaces}~\cite{MasterFace} examined the vulnerability of face recognition systems to dictionary attacks with reasonable success. However, the authors in~\cite{LimitMasterFace} pointed out the limited generalizability of face-based dictionary attacks across matchers, and contended that these attacks become less effective with increase in dimensionality of the facial representation. Dictionary attacks on speaker verification systems were introduced in~\cite{Speakerdict}. 

\textbf{Motivation:} Iris recognition is deployed in many applications due to its high accuracy and fast matching~\cite{AADHAR, Airport}. The \textit{biometric menagerie}~\cite{Menagerie} highlights that the \textit{wolf-like}~\cite{dodd} behavior of an individual (\textit{i.e.}, a single person fortuitously matches multiple people in a zero-effort imposter attack) is dominantly an image-specific issue stemming from non-ideal image acquisition in iris recognition systems, and not particularly a subject-specific issue. Iris image acquisition guidelines~\cite{IREXV} dictate specific requirements (iris radius $\geq$ 80 pixels, rotation $<15^{\circ}$ and daytime illumination). However, with commercial sensors mounted on hand-held devices, factors such as stand-off distances, indoor \textit{vs.} outdoor, and occlusions (\textit{e.g.}, eyeglasses) result in non-ideal conditions that can make iris templates extracted from such images vulnerable to false matches. The wolf-attack probability~\cite{WAP} can be also increased by using either a set of real or synthetic biometric samples that can adversarially match several users. We hypothesize that by carefully selecting users and further mixing their IrisCodes to form \textit{alpha-mixtures} can significantly increase the success of dictionary attack. In this work, we adopt two strategies to generate alpha-mixtures at template level, (i) combine wolves inherent in the population to generate alpha-wolves, and (ii) combine users selected via a search-based scheme (with or without wolves) to generate alpha-mammals.

\textbf{Contributions:} (a) We conduct a \textit{novel} vulnerability analysis of iris recognition systems that uses mixed IrisCodes, referred to as alpha-mixtures, to launch dictionary attacks at the template level. The mixture comprises alpha-wolves and alpha-mammals that are highly effective in matching arbitrary users not used in mixing. (b) We adopt different strategies for i) IrisCode selection (wolf selection and search optimization), ii) mixing of IrisCodes (using simple bitwise logical operators and CNN-based mixing) and iii) evaluation (log-Gabor and spatial Gabor encoding) on multiple datasets. (c) We examine the utility of synthetic IrisCodes as dictionary attacks against real IrisCodes. We demonstrate the effectiveness of our method under limited knowledge assumptions with cross-attacks.

\section{Related Work}
\label{Sec:Rel}
\subsection{Preliminaries}
\label{Sec:Prelim}
Iris recognition involves these steps~\cite{Introtobiom}. 1) Iris image \textit{acquisition} uses specialized sensors operating in the near-infrared spectrum (700-900nm). 2) Iris \textit{segmentation} extracts the colored annular region between the limbus and pupillary boundaries. 3) Iris \textit{encoding}, $\mathcal{E}(\cdot)$, obtains a compact template known as the IrisCode\footnote{Typically, the term ``IrisCode" has been associated with the Daugman-method~\cite{Iris1JD} for extracting iris templates; however, in this work, we use it to indicate any binary code extracted from the iris.}  using texture representation schemes such as Gabor filters. The IrisCode is typically a binary feature vector consisting of 2,048 bits that encodes the phase representation of the iris texture. Other types of encoding schemes have also been developed~\cite{LG,QSW,Local,DCT,KO}. 4) Iris \textit{matching}, $\mathcal{S}(\cdot,\cdot)$, uses fractional Hamming distance to measure the proportion of disagreement of the bits between two IrisCodes to produce a decision of match or non-match depending on the threshold, $\tau$, at a selected False Match Rate (FMR).

Daugman's IrisCode~\cite{Iris1JD} achieves extremely low FMR (1 in 26 million at HD=0.32) due to its high entropy~\cite{InfoIris} while maintaining fast matching using bitwise-Hamming distance. There has been a slew of other iris recognition methods using traditional filtering schemes, such as log-Gabor filters (LG)~\cite{LG}, spatial Gabor filters (QSW)~\cite{QSW}, local-intensity variations~\cite{Local}, DCT-based analysis~\cite{DCT}, cumulative sum based analysis~\cite{KO}. Deep learning based iris recognition systems perform end-to-end matching with comparable matching accuracy and are not limited to binary features or Hamming distance. Refer to~\cite{DLIris} for a survey of deep learning based iris recognition methods. Note that DL-based iris recognition typically do not use binary IrisCodes which is the focus of this work. In this work, we implement dictionary attacks using open-source implementation of \textit{lg} and \textit{qsw}-based features~\cite{USIT3}.

\subsection{Morphing \textit{vs}. Mixing} Erdogan~\cite{Gizem} proposed generating a dual-identity iris image by creating a composite of two irides using multiple strategies. Rathgeb and Busch~\cite{IrisCodeMorph} proposed \textit{morphing} two IrisCodes that matches the individuals whose IrisCodes contributed to the mixture. They performed random bit substitution, random row substitution and stability-based bit substitution to demonstrate that morphed IrisCodes can result in a fractional Hamming distance $<0.32$. Similarly,~\cite{IrisImageMorph} demonstrated that iris images from the left eye class can be morphed with images from the right eye class resulting in $>90\%$ successful attacks. Note that our method performs \textit{mixing} of IrisCodes using a function, $\mathcal{M}(\{IC_k\})$, where $k\geq2$. See the details of mixing in Sec.~\ref{Sec:Prop}. Unlike morphing, the mixed IrisCode can spoof multiple {\em other} identities, and not just the inputs to the mixing function.

       

\SetKwComment{Comment}{/* }{ */}


\section{Proposed Method}
\label{Sec:Prop}
Combining IrisCodes requires three inputs: (i) a set of seed IrisCodes, (ii) the number of seed IrisCodes to be combined, and (iii) a mixing function to combine the seed IrisCodes. We describe the inputs for the two methods developed in this work below. 
\subsection{Method I: Generating Alpha-wolves}
\label{Sec:wolfmix}
 We ideally want the alpha-mixture to behave as wolves that causes a high number of biometric collisions. A simple yet effective way of ensuring the wolf-like behavior of the mixture is to begin with a set of disjoint wolves as seed IrisCodes. The strategic selection of wolves as seed IrisCodes brings us to the concept of Doddington's zoo~\cite{dodd}. The biometric menagerie classifies those individuals as {\em wolves}  who successfully match other people (zero-effort imposter attack) resulting in high false matches. We utilize this phenomenon to rationalize our selection of wolves that will serve as the optimal set of seed IrisCodes. The \textit{best} number of wolf (seed) IrisCodes to be combined is a hyper-parameter determined during evaluation. We select a fixed set of seed IrisCodes that match imposters at a false match rate (\textit{e.g.}, FMR=0.01\%). Next, we combine wolves following \(\binom{n}{k}\), where $n$ denotes the number of wolves for a dataset (training set) $\mathcal{D}_{tr}$, $\lvert \mathcal{D}_{tr} \rvert = d$, and $k= \{2, \cdots, n\}$. We select bitwise operations, \textit{viz}., $AND (\&), OR (|)$ and $XOR (\oplus)$ operators for mixing IrisCodes. Therefore, we use the mixing function as follows: $\mathcal{M}_1(IC_1 \& IC_2); \mathcal{M}_1(IC_1 | IC_2); \mathcal{M}_1(IC_1 \oplus IC_2)$ for $k=2$. We select bitwise operators due to the binary nature of the IrisCode. Refer to Lines 1-14 in Algo.~\ref{alg:Merged}. The algorithmic time-complexity of wolf selection is $O(d)$ as it involves a single pass over the training set. The algorithmic time-complexity of wolf mixing is $O(kl)$, where $k*l << d$. We combine $k$ seed wolves using $l$ mixing functions resulting in $O(kl)$ time complexity for generating alpha-wolves. 

\subsection{Method II: Generating Alpha-mammals}
\label{Sec:search}
We hypothesize that while mixing wolves leads to ``alpha-wolves", more optimal combination of samples might exist in the dataset that does not necessarily involve wolves. We can consider a wolf as a point in the template space that is close to several other identities, simultaneously. It behaves as a centroid with maximal overlap with other identities. However, combining wolves may inadvertently push the mixed idenitity, \textit{i.e.}, the ``alpha-wolf", away from its optimal position, thereby reducing proximity with other individuals. This brings us to the idea of combining other members of the Doddington's zoo, such as mixing a wolf with a sheep (low False Match Rate and low False Non-Match Rate), that may lead to a more optimal dictionary attack. We \textit{search} for this optimal set of IrisCodes to be combined using a second methodology coined ``alpha-mammals". This hill-climbing approach makes no assumption about the need for multiple alpha-wolves nor on the number of IrisCodes that should be combined. We define the user coverage for an IrisCode $IC$ from the training set $\mathcal{D}_{tr}$ at a threshold $\tau$ as follows. 
\begin{equation}
    Cov(IC) = \sum_{i\in \mathcal{D}_{tr}} [\mathcal{S}(i,IC) \leq \tau]
    \label{eq:cov}
\end{equation}
\noindent The \textit{state space} is defined as the set of IrisCodes that are mixed together using some function $\mathcal{M}$. The \textit{action space} is defined as either the deletion or addition of any IrisCode to the current state space. Finally, the reward is defined as the user coverage described in Eqn.~\ref{eq:cov}. The algorithm starts with an empty set. In the first iteration, it always picks the wolf sample that leads to maximal coverage. Next, the algorithm optimizes over the set of users in $\mathcal{D}_{tr}$. Refer to Lines 15-46 in Algo.~\ref{alg:Merged}. The algorithmic time-complexity is $O(dtl)$, where $d$ is the size of the training set, $t$ is the number of iterations used and $l$ is the number of mixing functions. 

\begin{algorithm}[h]
\scriptsize
\caption{Generating Alpha-Mixtures}\label{alg:Merged}
\KwData{$\mathcal{D}_{tr}$, $\mathcal{E}(\cdot)$, $\mathcal{S}(\cdot,\cdot)$, $\tau$ and $\mathcal{M}(\{\cdot\})$}
\KwResult{Set of Alpha-wolves: $\alpha_{w}$, Alpha-mammals: $\alpha_{m}$ }

\textbf{Generating Alpha-wolves:}

\textit{Step I: Wolf selection} 

$\{IC_1,\cdots, IC_d\} \gets \mathcal{E}(\mathcal{D}_{tr})$, $\lvert \mathcal{D}_{tr} \rvert = d$ \Comment*[r]{ \footnotesize  IrisCodes} 
\If{$\mathcal{S}(IC_i,IC_j) \leq \tau$ $\forall j$ and $i \neq j$ \Comment*[r]{ \footnotesize check for false match}  } 
{ $\mathcal{W} \gets IC_i$\Comment*[r]{\footnotesize seed wolves}  
}
\textit{Step II: Wolf mixing} \Comment*[r]{$\lvert \mathcal{W} \rvert = n$}  
\For{$k=2$ \KwTo $n$}
{
$\mathcal{C}_k \gets \binom{n}{k};$   
$\mathcal{P}_k \gets \mathcal{W}\{\mathcal{C}_k\}$ \Comment*[r]{ \footnotesize select combinations of seed wolves}
\For{$l=1$ \KwTo $3$}
{
$\alpha_{w_{kl}} \gets \mathcal{M}_l(\{\mathcal{P}_k\})$ \Comment*[r]{ \footnotesize  mix seed wolves with 3 bitwise ops.} 
}
}

\textbf{Generating Alpha-mammals:}

$q \leftarrow \phi$ \Comment*[r]{ \footnotesize initialize an empty IC set} 
$c \leftarrow 0 $ \Comment*[r]{ \footnotesize set current coverage to 0}
$c_n \leftarrow 0 $ \Comment*[r]{\footnotesize set best neighbor coverage to 0}





\For{$l=1$ \KwTo $3$}{

\While{ $c_n \geq c$ }{

$c\leftarrow c_n$
$q \leftarrow  q_n$

\textit{Step I: Check if appending sample helps.} \;
\For{$k=1$ \KwTo $|\mathcal{D}_{tr}|$}
{
$q' \leftarrow q.append(IC_k)$

$IC \leftarrow \mathcal{M}_l(q')$ \Comment*[r]{\footnotesize mix set of ICs}

$c_k$ = $\sum_{i\in \mathcal{D}_{tr}} [\mathcal{S}(i,IC) \leq \tau]$ \Comment*[r]{\footnotesize cov on new IC}

\If{$c_k \geq c_n$ \Comment*[r]{\footnotesize cov improved, update cov and IC set}}  
{
$c_n \leftarrow c_k$

$q_n \leftarrow q.append(IC_k)$

}
}

\textit{Step II: Check if removing sample helps.} \;
\For{$k=1$ \KwTo $|q|$}
{

$q' \leftarrow q.remove(IC_k)$

$IC \leftarrow \mathcal{M}_l(q')$ \Comment*[r]{\footnotesize mix set of ICs}

$c_k$ = $\sum_{i\in \mathcal{D}} [\mathcal{S}(i,IC) \leq \tau]$ \Comment*[r]{\footnotesize cov on new IC}

\If{$c_k \geq c_n$ \Comment*[r]{\footnotesize cov improved, update cov and IC set}}
{
$c_n \leftarrow c_k$

$q_n \leftarrow q.remove(IC_k)$

}
}
}
$\alpha_{m_l} \leftarrow \mathcal{M}_l(q)$
}

\textbf{Return:} $\alpha_{w}$,$\alpha_{m}$   \Comment*[r]{\footnotesize alpha-mixtures} 
\end{algorithm}
\section{Experimental Design}

We use two real iris datasets, namely, \textbf{IITD}~\cite{IITD}, and \textbf{CASIA-IrisV4-Thousand}~\cite{CASIAv4} and a synthetic iris dataset, namely, \textbf{CASIA-IrisV4-Synthetic}~\cite{CASIAv4}. We consider left and right eye images as separate classes for both real datasets. The synthetic dataset has only a single eye for each identity. The IITD dataset comprises 224 subjects with 1,188 left eye images and 1,052 right eye images that is used for both wolf selection and evaluation. We use USIT v3.0.0 toolkit\footnote{We use the Windows executable provided by the original authors at \url{https://www.wavelab.at/sources/USIT/} in Method I and custom built Linux executable in Method II.} to perform iris segmentation (using contrast-adjusted Hough transform (\textit{caht})), IrisCode feature extraction (using log-Gabor (\textit{lg}) and quadrature spline wavelet (\textit{qsw})), and matching (using fractional Hamming distance (\textit{hd})). Note that USIT v3.0 converts the binary IrisCode to an 8-bit unsigned integer in [0, 255] and then flattens it to produce a 1-D vector of $1,280$ bytes ($2^3\times2^7\times10$ equivalent to $20\times512$ $= 10,240$ bits). Although the original authors indicated possible correlations between the first and last 10 rows~\cite{IrisCodeMorph}, we use the entire template. We perform matching using the 1-D flattened representation while mixing is performed on the binary valued 2-D template.
\begin{figure}
     \centering
     \begin{subfigure}[b]{0.12\textwidth}
         \centering
         \includegraphics[width=\textwidth]{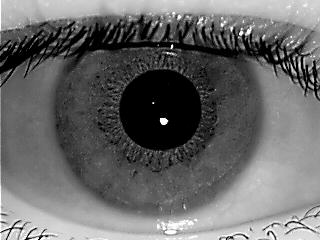}
     \end{subfigure}
     \begin{subfigure}[b]{0.12\textwidth}
         \centering
         \includegraphics[width=\textwidth]{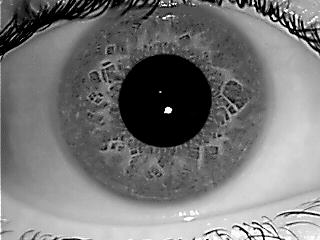}
     \end{subfigure}
     \begin{subfigure}[b]{0.12\textwidth}
         \centering
         \includegraphics[width=\textwidth]{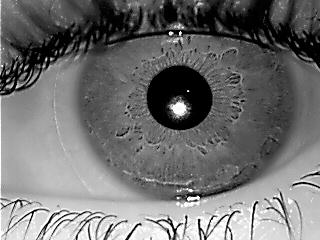}
     \end{subfigure} \\
     \begin{subfigure}[b]{0.12\textwidth}
         \centering
         \includegraphics[width=\textwidth]{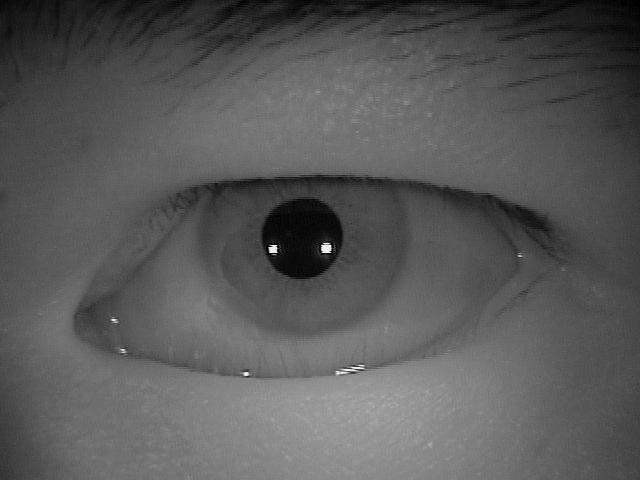}
     \end{subfigure}
     \begin{subfigure}[b]{0.12\textwidth}
         \centering
         \includegraphics[width=\textwidth]{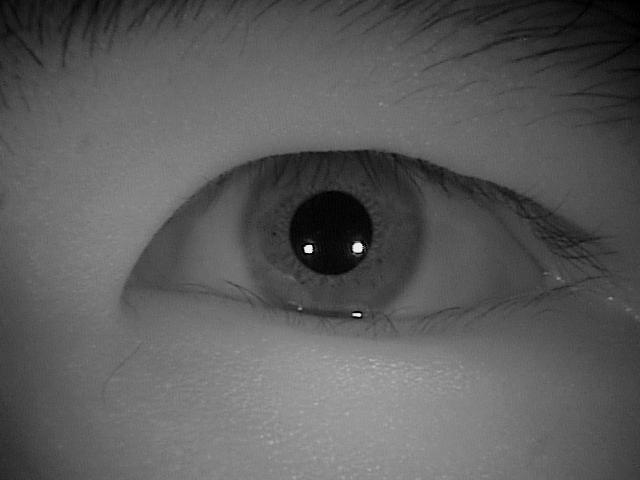}
     \end{subfigure}
     \begin{subfigure}[b]{0.12\textwidth}
         \centering
         \includegraphics[width=\textwidth]{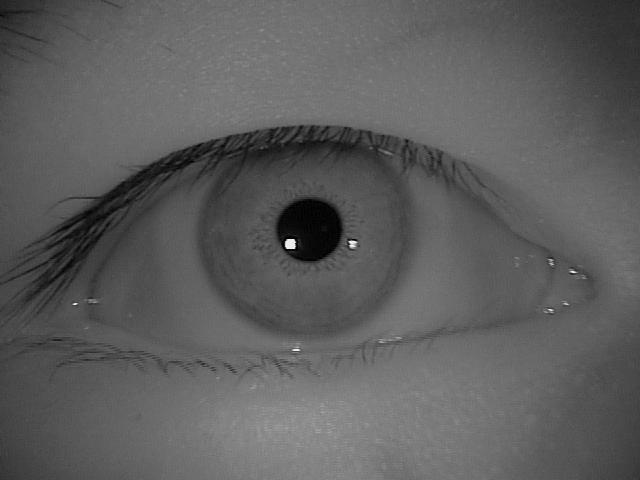}
     \end{subfigure} 
     \begin{subfigure}[b]{0.12\textwidth}
         \centering
         \includegraphics[width=\textwidth]{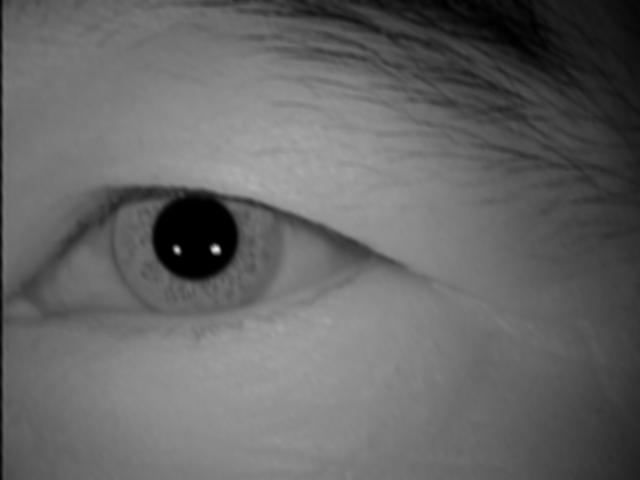}
     \end{subfigure}
     \begin{subfigure}[b]{0.12\textwidth}
         \centering
         \includegraphics[width=\textwidth]{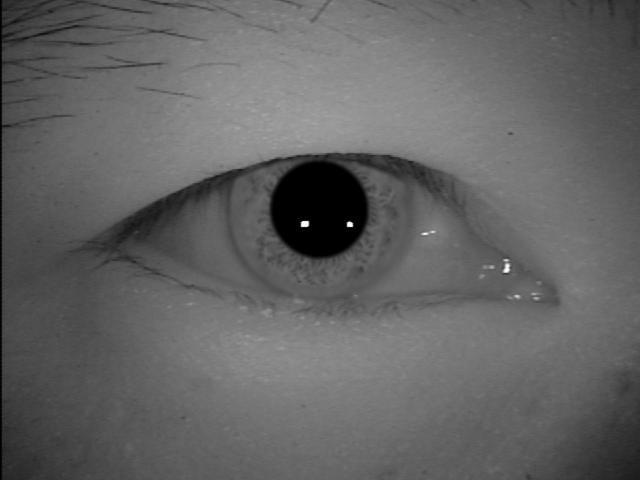}
     \end{subfigure}
     \begin{subfigure}[b]{0.12\textwidth}
         \centering
         \includegraphics[width=\textwidth]{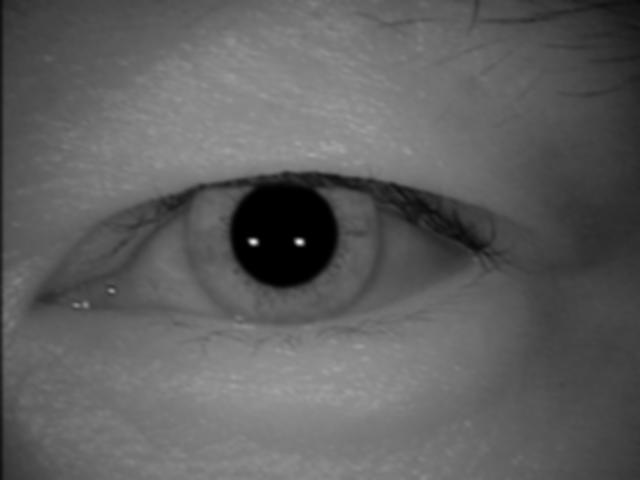}
     \end{subfigure}
        \caption{Examples of wolves used in generating alpha-wolves belonging to IITD (top row), CASIA-IrisV4-Thousand (middle row) and CASIA-IrisV4-Synthetic (bottom row) datasets. Note that, visually, wolves are high quality iris images.}  %
   \label{Fig:wolfexample}  
\end{figure}

CASIA-IrisV4-Thousand comprises 1,000 subjects with 10,000 left eye images and 10,000 right eye images. We use the COTS Neurotechnology VeriEye 12.4 SDK for segmentation, Libor Masek code for texture and mask generation, and USIT v3.0 for encoding and matching. We discard inadmissible IrisCodes upon visual inspection (\textit{e.g.}, all 0's). Next, we select wolves in 1:10 and 3:10 ratio from the entire set of 1,000 identities to form the training set, resulting in first 93 subjects using $lg$ @FMR=0.05\%, and the first 265 subjects using $qsw$ features @FMR=0.01\% (some identities were manually discarded due to unreliable encoding). We adopt this strategy to examine if wolves selected from a subset of the population (training set) can be effective against unseen subjects from the same dataset (test set). This helps us evaluate the effectiveness of the method with access to limited number of wolf samples. In this case, the user coverage is evaluated on the remainder of the target population (test set) excluding the subset from which the wolves were selected. See examples of wolves in Fig.~\ref{Fig:wolfexample}. For computing the alpha-mammals, we search across the first 93 users and test on the remaining set of 907 users on the CASIA-IrisV4-Thousand dataset for both the $qsw$ and $lg$ features. 

\section{Results and Analysis}
\label{Sec: Results}

\subsection{Results for Method I: Alpha-wolves}
\label{Sec:ResultsMethodI}

We select that mixed IrisCode that produces the highest false matches across the test set as the {\em best} alpha-wolf, \textit{i.e.}, $\alpha_{w,best}= \arg \max_{\alpha_{w_{k}}} \mathcal{S}(\alpha_{w_{k}},IC_q); \forall q \in \mathcal{D}_{te}$. Here, $\alpha_{w_{k}}$ denotes the set of alpha-wolves corresponding to mixing $k$ out of $n$ seed wolves, $\mathcal{S}(\cdot, \cdot)$ is the iris matcher, and $\mathcal{D}_{te}$ denotes the test set. We report the proportion of identities where at least one sample matched with the \textit{best} alpha-wolf generated using different bitwise operators (AND, OR, XOR) and different feature extraction schemes (\textit{lg} and \textit{qsw}) on both eyes (left and right). We refer to this proportion as the user coverage that quantifies the success of the dictionary attack. We observe that as the number of wolves increase, the OR-mixture results in denser codes, while the AND-mixture results in sparser codes and the XOR-mixture contains equal proportion of 1's and 0's in the alpha-wolves. So, we restrict to mixing two, three and four seed wolves. 
\begin{table}[h]
\centering
\caption{Alpha-wolf attacks on IITD $lg$-Left dataset.}
\scalebox{0.85}{
\begin{tabular}{ccccc} \hline
\multicolumn{1}{c}{\multirow{3}{*}{\begin{tabular}[c]{@{}c@{}}\# seed \\ wolves\end{tabular}}} & \multicolumn{4}{c}{\begin{tabular}[c]{@{}c@{}}Proportion of Users (\%) Covered \\ @ False Match Rate for OR/AND/XOR\end{tabular}} \\ \cline{2-5}
\multicolumn{1}{c}{}  & 0.001\%                       & 0.01\%                        & 0.1\%                               & 1\%                                 \\ \hline
2     &  0.9/0.9/\textbf{10.7}               &    0.9/0.9/\textbf{20.5}               &   1.8/1.8/\textbf{79.0}                   & 2.7/3.1/\textbf{89.7}                  \\
3     &   0.0/0.0/0.0             &   0.0/0.0/0.9           &   2.2/2.2/0.9                &   1.8/3.1/0.4           \\
4    &   0.0/0.0/5.8              &    0.0/0.0/6.2              & 1.8/1.8/50.4                        & 0.9/3.1/64.3       \\ \hline              
\end{tabular}}
\label{Tab:1real}
\end{table}

\begin{table}[h]
\centering
\caption{Alpha-wolf attacks on IITD $lg$-Right dataset.}
\scalebox{0.85}{
\begin{tabular}{ccccc} \hline
\multicolumn{1}{c}{\multirow{3}{*}{\begin{tabular}[c]{@{}c@{}}\# seed\\ wolves\end{tabular}}} & \multicolumn{4}{c}{\begin{tabular}[c]{@{}c@{}}Proportion of Users (\%) Covered \\ @ False Match Rate for OR/AND/XOR\end{tabular}} \\ \cline{2-5}
\multicolumn{1}{c}{}          & 0.001\%                       & 0.01\%                        & 0.1\%                               & 1\%                                 \\ \hline
2               & 0.9/0.9/\textbf{17.9}                   & 0.9/0.9/\textbf{21.4}                    &  2.2/1.8/\textbf{82.1}                    & 2.7/4.0/84.4                   \\
3              & 0.0/0.0/0.0           & 0.0/0.0/0.0                 & 2.2/2.7/0.9      &  3.1/3.6/0.9      \\
4              & 0.0/0.0/15.6                   &   0.0/0.0/17.4                  & 2.2/1.8/79.9              & 2.7/3.1/\textbf{88.3}       \\ \hline              
\end{tabular}}
\label{Tab:2real}
\end{table}

\begin{table}[h]
\centering
\caption{Alpha-wolf attacks on IITD $qsw$-Left dataset.}
\scalebox{0.85}{
\begin{tabular}{ccccc} \hline
\multicolumn{1}{c}{\multirow{3}{*}{\begin{tabular}[c]{@{}c@{}}\# seed \\ wolves\end{tabular}}} & \multicolumn{4}{c}{\begin{tabular}[c]{@{}c@{}}Proportion of Users (\%) Covered \\ @ False Match Rate for OR/AND/XOR\end{tabular}} \\ \cline{2-5}
\multicolumn{1}{c}{}          & 0.001\%                 & 0.01\%                   & 0.1\%                         & 1\%                                 \\ \hline
2     &  0.9/0.9/\textbf{20.9}              & 0.9/0.9/\textbf{25.0}               & 2.2/2.2/\textbf{91.5}                       &  4.5/3.6/\textbf{96.4}              \\
3     & 0.0/0.0/0.0    &   0.0/0.0/0.4                 &  3.1/2.7/1.3                 & 4.5/3.6/1.8              \\
4    &  0.0/0.0/4.0              &   0.0/0.0/5.4                &  2.7/3.1/83.0                       & 4.0/2.2/94.2      \\ \hline              
\end{tabular}}
\label{Tab:3real}
\end{table}

\begin{table}[h]
\centering
\caption{Alpha-wolf attacks on IITD $qsw$-Right dataset.}
\scalebox{0.85}{
\begin{tabular}{ccccc} \hline
\multicolumn{1}{c}{\multirow{3}{*}{\begin{tabular}[c]{@{}c@{}}\# seed\\ wolves\end{tabular}}} & \multicolumn{4}{c}{\begin{tabular}[c]{@{}c@{}}Proportion of Users (\%) Covered \\ @ False Match Rate for OR/AND/XOR\end{tabular}} \\ \cline{2-5}
\multicolumn{1}{c}{}        & 0.001\%                  & 0.01\%                  & 0.1\%                 & 1\%                                 \\ \hline
2     & 0.9/0.9/\textbf{31.7}                & 0.9/0.9/\textbf{42.4}               &  2.2/2.7/\textbf{88.8}                   &  4.5/3.6/\textbf{90.6}                 \\
3     & 0.0/0.0/0.0          &   0.0/0.4/0.0                &   2.7/3.1/0.9               &  4.0/4.0/0.9            \\
4    & 0.0/0.4/13.4               & 0.0/0.4/16.9                  & 2.2/2.7/88.8                     &  3.6/4.0/90.1      \\ \hline              
\end{tabular}}
\label{Tab:4real}
\end{table}

We present results on the IITD dataset in Tables~\ref{Tab:1real},\ref{Tab:2real},\ref{Tab:3real} and \ref{Tab:4real}. We observe that the overall user coverage is the highest for the XOR operator for mixing IrisCodes. We report the results at four $FMR (\%)$ values$=\{0.001, 0.01, 0.1, 1\}$ on IITD. As observed in the results, \textit{qsw} feature-based IrisCode is more successful in generating dictionary attacks compared to log-Gabor (\textit{lg}) feature-based IrisCode. 
Our attack achieves up to 31.7\% user coverage @FMR=0.001\% by XOR-ing two identities on \textit{qsw}-right, 42.4\% user coverage @FMR=0.01\% by XOR-ing two identities on \textit{qsw}-right, 91.5\% user coverage @FMR=0.1\% by XOR-ing two identities on \textit{qsw}-left dataset, and up to 96.4\% user coverage @FMR=1\% by XOR-ing two identities on \textit{qsw}-left. 

\begin{table}[h]
\centering
\caption{Alpha-wolf attacks on CASV4-Th $lg$-Left dataset.}
\scalebox{0.86}{
\begin{tabular}{ccccc} \hline
\multicolumn{1}{c}{\multirow{3}{*}{\begin{tabular}[c]{@{}c@{}}\# seed \\ wolves\end{tabular}}} & \multicolumn{4}{c}{\begin{tabular}[c]{@{}c@{}}Proportion of Users (\%) Covered \\ @ False Match Rate for OR/AND/XOR\end{tabular}} \\ \cline{2-5}
\multicolumn{1}{c}{}  & 0.01\%                       & 0.05\%                        & 0.1\%                               & 1\%                                 \\ \hline
2     &  0.0/2.2/0.0             &   0.0/\textbf{57.1}/1.3               &  0.3/\textbf{64.4}/3.9                  & 33.6/\textbf{98.3}/68.0                 \\
3     &   0.0/\textbf{2.4}/0.9          &   0.0/56.5/2.2         &   0.0/63.9/2.3                &   1.0/97.7/6.6           \\
4    &   0.0/2.4/0.0             &    0.0/55.8/0.7             &  0.0/62.5/0.7                      & 0.0/97.2/3.1      \\ \hline              
\end{tabular}}
\label{Tab:1realCASIAv4}
\end{table}

\begin{table}[h]
\centering
\caption{Alpha-wolf attacks on CASV4-Th $lg$-Right dataset.}
\scalebox{0.85}{
\begin{tabular}{ccccc} \hline
\multicolumn{1}{c}{\multirow{3}{*}{\begin{tabular}[c]{@{}c@{}}\# seed \\ wolves\end{tabular}}} & \multicolumn{4}{c}{\begin{tabular}[c]{@{}c@{}}Proportion of Users (\%) Covered \\ @ False Match Rate for OR/AND/XOR\end{tabular}} \\ \cline{2-5}
\multicolumn{1}{c}{}  & 0.01\%                       & 0.05\%                        & 0.1\%                               & 1\%                                 \\ \hline
2     &  0.3/1.0/\textbf{10.0}          & 0.4/2.7/\textbf{67.9}            & 0.6/4.2/\textbf{70.3}          &  3.2/25.3/\textbf{93.7}             \\
3     &   0.2/1.0/0.2           & 0.3/3.3/1.4                              & 0.3/4.2/1.8                    &   1.0/15.8/2.9       \\
4    &  0.2/0.5/2.2             & 0.2/3.0/45.7                             & 0.2/4.1/51.0                   & 0.5/12.4/78.6      \\ \hline              
\end{tabular}}
\label{Tab:2realCASIAv4}
\end{table}

\begin{table}[]
\centering
\caption{Alpha-wolf attacks on $qsw$-CASV4-Th Left dataset.}
\scalebox{0.87}{
\begin{tabular}{ccccc} \hline
\multicolumn{1}{c}{\multirow{3}{*}{\begin{tabular}[c]{@{}c@{}}\# seed \\ wolves\end{tabular}}} & \multicolumn{4}{c}{\begin{tabular}[c]{@{}c@{}}Proportion of Users (\%) Covered \\ @ False Match Rate for OR/AND/XOR\end{tabular}} \\ \cline{2-5}
\multicolumn{1}{c}{}  & 0.001\%                       & 0.01\%                        & 0.1\%                               & 1\%                                 \\ \hline
2     & 0.0/0.3/\textbf{0.5}           & 0.0/0.8/\textbf{2.4}            & 0.8/2.7/\textbf{26.4}           & 54.0/32.8/\textbf{99.5}        \\
3     & 0.0/0.3/0.0                    & 0.0/0.7/0.0                     & 0.5/2.5/1.4                      &  32.8/25.4/8.4     \\
4    & 0.0/0.3/0.0                     & 0.0/0.7/1.7                     &  0.0/2.3/5.6                     &  10.6/18.5/77.7    \\ \hline              
\end{tabular}}
\label{Tab:3realCASIAv4}
\end{table}

\begin{table}[]
\centering
\caption{Alpha-wolf attacks on $qsw$-CASV4-Th Right dataset.}
\scalebox{0.84}{
\begin{tabular}{ccccc} \hline
\multicolumn{1}{c}{\multirow{3}{*}{\begin{tabular}[c]{@{}c@{}}\# seed \\ wolves\end{tabular}}} & \multicolumn{4}{c}{\begin{tabular}[c]{@{}c@{}}Proportion of Users (\%) Covered \\ @ False Match Rate for OR/AND/XOR\end{tabular}} \\ \cline{2-5}
\multicolumn{1}{c}{}  & 0.001\%                       & 0.01\%                        & 0.1\%                               & 1\%                                 \\ \hline
2     &   0.0/6.3/0.0              & 0.3/\textbf{72.8}/0.5                &  10.7/\textbf{97.3}/10.6                & 73.1/\textbf{99.3}/88.6             \\
3     & 0.0/\textbf{9.5}/0.0                & 0.1/72.7/0.6                & 2.4/97.2/0.7                          & 45.9/99.3/73.8         \\
4    &  0.0/9.5/0.0              &  0.1/61.7/0.3                          &  0.4/93.2/1.2                         & 11.6/97.8/24.7     \\ \hline              
\end{tabular}}
\label{Tab:4realCASIAv4}
\end{table}

We present results on CASIA-IrisV4-Thousand (for brevity we will refer it as CASV4-Th) dataset at $FMR(\%)=\{0.01, 0.05, 0.1, 1\}$ for $lg$-feature in Tables~\ref{Tab:1realCASIAv4} and~\ref{Tab:2realCASIAv4}, and at $FMR(\%)=\{0.001, 0.01, 0.1, 1\}$ for $qsw$-feature in Tables~\ref{Tab:3realCASIAv4} and~\ref{Tab:4realCASIAv4}. This is because we observe inherently poor performance on the CASV4-Th dataset, so we selected wolves at FMR=0.05\% when applicable. Alpha-wolves achieve upto 9.5\% user coverage @FMR=0.001\% by AND-ing three identities using \textit{qsw }feature on the CASV4-Th Right dataset, 72.8\% user coverage @FMR=0.01\% by AND-ing two identities using \textit{qsw} features on the CASV4-Th Right dataset, 67.9\% @FMR=0.05\% by XOR-ing two identities using \textit{lg} features on the CASV4-Th Right dataset, 97.3\% user coverage @FMR=0.1\% by AND-ing two identities using \textit{qsw }feature on the CASV4-Th Right dataset, and upto 99.5\% user coverage @FMR=1\% by XOR-ing two identities using \textit{qsw} features on the CASV4-Th Left dataset.

\subsection{Results for Method II: Alpha-mammals}

\begin{table}[]
\centering
\caption{Alpha-mammal attacks on the IITD dataset. }
\scalebox{0.84}{
\begin{tabular}{cccc} \hline
\multicolumn{1}{c}{\multirow{3}{*}{feature-laterality}} & \multicolumn{3}{c}{\begin{tabular}[c]{@{}c@{}}Proportion of Users (\%) Covered \\ @ False Match Rate for OR/AND/XOR\end{tabular}} \\ \cline{2-4}
\multicolumn{1}{c}{}                 & 0.01\%                  & 0.1\%                 & 1\%                                 \\ \hline
$lg$-Left            &         0.9/0.9/51.6      &           28.7/57.8/28.7         &         53.8/76.7/53.8         \\
$lg$-Right     &   0.9/0.9/4.1               &   4.1/4.1/85.1             &   11.3/12.6/90.1       \\
$qsw$-Left         &       0.9/0.9/\textbf{59.6}      &   58.7/58.7/64.1   &   88.3/88.3/\textbf{92.4}           \\
$qsw$-Right        &   0.9/0.9/6.3          &    4.1/4.5/\textbf{90.5}    &    13.5/13.5/92.3    \\ \hline             
\end{tabular}}
\label{Tab:search_IITD}
\end{table}

\begin{table}[]
\centering
\caption{Alpha-mammal attacks on the CASV4-Th dataset. Here, we discard mixtures with more than 70\% of 1's or 0's. }
\scalebox{0.80}{
\begin{tabular}{cccc} \hline
\multicolumn{1}{c}{\multirow{3}{*}{feature-laterality}} & \multicolumn{3}{c}{\begin{tabular}[c]{@{}c@{}}Proportion of Users (\%) Covered \\ @ False Match Rate for OR/AND/XOR\end{tabular}} \\ \cline{2-4}
\multicolumn{1}{c}{}                 & 0.01\%                  & 0.1\%                 & 1\%                                 \\ \hline
$lg$-Left      &  4.8/\textbf{59.5}/15.6 &  6.1/\textbf{92.6}/23.6 &  46.5/99.0/92.7 \\
$lg$-Right     &  4.6/6.2/51.8 &  12.0/19.3/76.6 & 48.0/75.5/98.6 \\
$qsw$-Left     & 0.1/1.6/2.2 & 3.2/5.1/28.1 & 83.2/80.3/99.1  \\
$qsw$-Right    & 0.4/0.9/74.3 & 2.2/4.8/71.1 & 71.1/71.1/\textbf{99.5}\\ \hline           
\end{tabular}}
\label{Tab:search_casia}
\end{table}

\begin{figure}[h]
     \centering
     \begin{subfigure}[b]{\columnwidth}
         \centering
         \includegraphics[width=\textwidth]{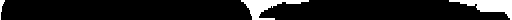}
         \caption{alpha-mammal IrisCode mask}
         
     \end{subfigure}
     \begin{subfigure}[b]{\columnwidth}
         \centering
         \includegraphics[width=\textwidth]{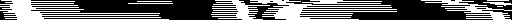}
         \caption{alpha-mammal IrisCode}
        
     \end{subfigure}
    
    \caption{Example of an alpha-mammal computed on the IITD dataset wherein a majority of the IrisCode is covered by the mask.} 
    \label{Fig:alphamammal}
\end{figure}

In this section, we present our results on dictionary attacks using the ``alpha-mammal" hill climbing algorithm. In Table \ref{Tab:search_IITD}, we present the results obtained on the IITD dataset for the $lg$ and $qsw$ features at $FMR(\%)=\{0.01, 0.1, 1.0\}$. Similarly, we show the results on the CASV4-Th dataset in Table \ref{Tab:search_casia}. It is also important to note that, unlike alpha-wolves, we only obtain one alpha-mammal per search process. Thus this attack is always one-shot. 
Overall, we saw that for the IITD dataset, the XOR operator has the best performance except for the $lg$ feature and the left eye class. In this specific exception, the AND operator outperforms the XOR operation at $FMR(\%)=\{0.1, 1\}$. We saw some interesting mixtures, especially for the AND operator and $lg$ feature on left IrisCodes, where the mixed IrisCode mask ends up covering a majority of the eye region. This may indicate that specific regions in the alpha-mammal IrisCode that are not occluded contribute significantly towards a successful dictionary attack. We show a specific example of such a scenario in Fig.~\ref{Fig:alphamammal}. Therefore, after more than three IrisCodes are combined we limit the lateral movement, \textit{i.e.}, unless we see an improvement in the reward function we terminate the search. This reduces the number of IrisCodes that are combined and thus limits the occluded region.





\begin{figure}
     \centering
     \begin{subfigure}[]{0.28\textwidth}
         \centering
         \includegraphics[width=\textwidth]{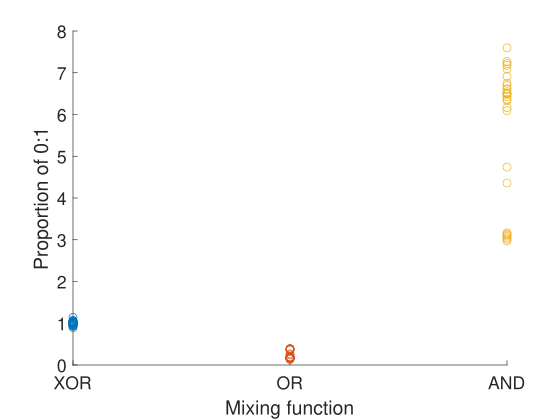}
     \end{subfigure} 
     \begin{subfigure}[]{0.28\textwidth}
         \centering
         \includegraphics[width=\textwidth]{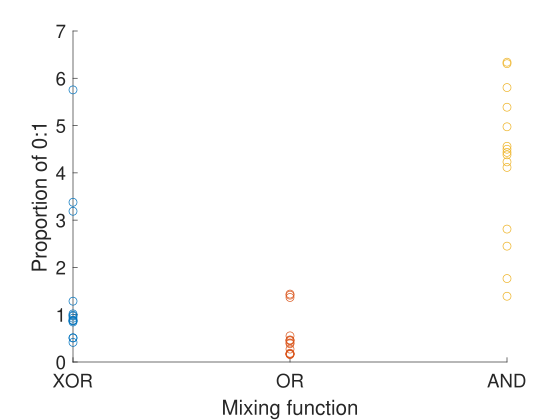}  
     \end{subfigure} 
        \caption{Distribution of 1's and 0's of alpha-wolves using two wolves for IITD $lg$-Right (left) and CASV4-Th $qsw$-Right (right).}
    \label{Fig:distribution} 
\end{figure}

\section{Discussion}
\label{Sec:disc}
\subsection{Analyzing alpha-mixtures}

 We discuss our findings and offer insights into the behavior of the \textit{alpha-mixtures}. Surprisingly, we observe that the seed IrisCodes that generate the alpha-mixtures belong to good quality ocular images with little or no occlusion, thereby alleviating concerns regarding fragile bits~\cite{bestbits}.
 We further observe that increasing the number of wolves/mammals in the mixture does not necessarily increase the chances of higher user coverage. In a majority of cases, mixing two wolves/mammals yields the best user coverage. In terms of mixing operators, XOR appears to produce the highest user coverage in most of the cases. XOR performs logical inequality that produces True/1 only if both bits disagree, implying XOR-mixing inherently combines highly dissimilar wolves, while OR- and AND-mixing combines relatively similar wolves. We use Normalized Compression Distance (NCD) to examine how `similar' or `dissimilar' the alpha-wolves ($\alpha_W$) are compared to the seed wolves ($W$). NCD computes the similarity between objects by evaluating the binary size of their compressed versions, $C(\cdot)$. $NCD(W, \alpha_{W})= \frac{C(W\alpha_{W}) - min\{C(W),C(\alpha_{W})\}}{max\{C(W),C(\alpha_{W})\}}$, $0<NCD<1$. We observe that NCD is highest for XOR-mixed alpha-wolves ($\approx0.98$) compared to AND- and OR-mixed alpha-wolves ($\approx0.85$) on the IITD dataset.\\
\hspace{0.4cm} Alternatively, in the context of information theory, Daugman suggests that there should be no uncertainty about the identity $X$ denoted by a biometric  signal $Y$, \textit{i.e.}, $H(X|Y)=0$~\cite{InfoIris}. However, $Y_{alphamix}$ combines biometric signals from two identities (say, $X_1$ and $X_2$), thus increasing the conditional entropy, $H(X_1,X_2|Y_{alphamix}) = H(X_1|Y_{alphamix}) + H(X_2|Y_{alphamix}, X_1) = H(X_2|Y_{alphamix}) + H(X_1|Y_{alphamix}, X_2)$. We speculate that higher conditional entropy in the alpha-mixtures may be responsible for increase in biometric collisions, thereby producing higher false matches.

\begin{figure}[h]
     \centering
     \begin{subfigure}[b]{\columnwidth}
         \centering
         \includegraphics[width=\textwidth]{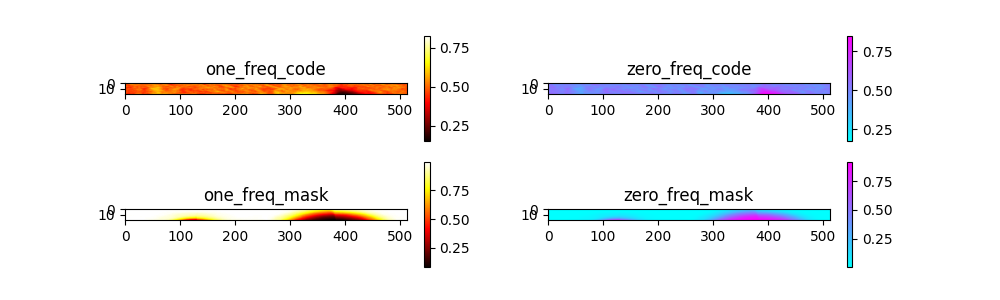}
         \caption{The heatmap of IITD left eyes with \textit{lg} feature.}
         
     \end{subfigure}
     \begin{subfigure}[b]{\columnwidth}
         \centering
         \includegraphics[width=\textwidth]{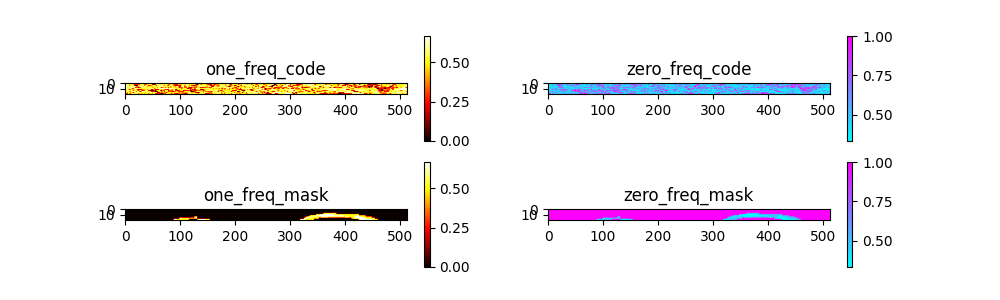}
         \caption{The heatmap of alpha-wolf of IITD left eyes with \textit{lg} feature.}
        
     \end{subfigure}

    \caption{Examples of the heatmap of `0's and `1's frequency. } 
    \label{Fig:heatmap}
\end{figure}

We investigate the distribution of 1's and 0's to understand the behavior of the \textit{mixing function}. Fig.~\ref{Fig:distribution} analyzes the proportion of 0's to 1's in the alpha-wolves from the IITD right-$lg$ feature and from CASV4-Th right-$qsw$-based IrisCodes. It is surprising to note that in cases where the XOR mixing function yields the highest coverage (IITD right-$lg$), the alpha-wolves are tightly clustered around one, indicating equal proportion of 1's and 0's. However, whenever the AND-mixing function produces higher user coverage compared to XOR (CASV4-Th right-$qsw$), we observe the proportion of 0's to 1's in XOR-ed alpha-wolves to be scattered. We present a statistical analysis (mean and standard deviation) of the bits of the IrisCodes and their masks for the original IrisCodes, seed wolves and alpha-wolves in Table~\ref{Tab:Disstats}. We observe that the XOR-mixed IrisCodes are consistent with the original IrisCodes, but the IrisCode masks become sparser, \textit{i.e.}, contain more 0's. We qualitatively analyze the frequency of `1' and `0' bits in original and alpha-wolf IrisCodes and masks in Fig.~\ref{Fig:heatmap}.


\begin{table}[]
 \centering
 \caption{Statistics of bit `1' in the original, seed codes and XOR-mixed alpha-wolves from the IITD and CASV4-Th datasets for IrisCodes and their masks.}
 \scalebox{0.82}{
 \begin{tabular}{cl|ll||ll|}
 \cline{3-6}
 \multicolumn{2}{c}{\begin{tabular}[c]{@{}c@{}}\\ \end{tabular}} & \multicolumn{2}{|c||}{\begin{tabular}[c]{@{}c@{}} IITD\\ \end{tabular}} & \multicolumn{2}{c|}{\begin{tabular}[c]{@{}c@{}} CASV4-Th\\ \end{tabular}}\\
 \cline{3-6} 
 & \multicolumn{1}{c|}{\begin{tabular}[c]{@{}c@{}}feature- \\ laterality\end{tabular}} & \multicolumn{1}{c|}{\begin{tabular}[c]{@{}c@{}}code \\ $\mu\pm\sigma$\end{tabular}} & \multicolumn{1}{|c||}{\begin{tabular}[c]{@{}c@{}}mask \\ $\mu\pm\sigma$\end{tabular}} & \multicolumn{1}{c|}{\begin{tabular}[c]{@{}c@{}}code \\ $\mu\pm\sigma$\end{tabular}} & \multicolumn{1}{|c|}{\begin{tabular}[|c]{@{}c@{}}mask \\ $\mu\pm\sigma$\end{tabular}}\\
 \hline
 \parbox[t]{2mm}{\multirow{4}{*}{\rotatebox[origin=c]{90}{original}}} & \textit{lg}-left &0.49$\pm$0.03&0.86$\pm$0.11&0.5$\pm$0.03&0.9$\pm$0.16\\
 & \textit{lg}-right &0.49$\pm$0.05&0.85$\pm$0.17&0.48$\pm$0.05&0.77$\pm$0.2\\
 & \textit{qsw}-left &0.55$\pm$0.5&0.83$\pm$0.17&0.56$\pm$0.07&0.9$\pm$0.16\\
 & \textit{qsw}-right &0.54$\pm$0.06&0.84$\pm$0.19&0.59$\pm$0.09&0.77$\pm$0.2\\
 \hline \hline 
 \parbox[t]{2mm}{\multirow{4}{*}{\rotatebox[origin=c]{90}{seed codes}}} & \textit{lg}-left &0.49$\pm$0.01&0.92$\pm$0.03&0.35$\pm$0.12&0.36$\pm$0.29\\
 & \textit{lg}-right &0.49$\pm$0.01&0.85$\pm$0.09&0.47$\pm$0.04&0.64$\pm$0.24\\
 & \textit{qsw}-left &0.49$\pm$0.005&0.86$\pm$0.10&0.45$\pm$0.16&0.54$\pm$0.31\\
 & \textit{qsw}-right &0.56$\pm$0.03&0.85$\pm$0.09&0.68$\pm$0.23&0.22$\pm$0.18\\
 \hline \hline 
\parbox[t]{2mm}{\multirow{4}{*}{\rotatebox[origin=c]{90}{$\alpha_{wolves}$}}} & \textit{lg}-left&0.45$\pm$0.1&0.06$\pm$0.01&0.44$\pm$0.09&0.46$\pm$0.21\\
 & \textit{lg}-right &0.47$\pm$0.05&0.13$\pm$0.07&0.48$\pm$0.03&0.31$\pm$0.2\\
 & \textit{qsw}-left &0.49$\pm$0.05&0.14$\pm$0.09&0.48$\pm$0.16&0.4$\pm$0.25\\
 & \textit{qsw}-right &0.46$\pm$0.03&0.13$\pm$0.07&0.46$\pm$0.18&0.31$\pm$0.14\\
 \hline 
 \end{tabular}}
 \label{Tab:Disstats}
 \end{table}

\begin{figure}[h]
     \centering
     \begin{subfigure}[b]{0.12\textwidth}
         \centering
         \includegraphics[width=\textwidth, height=0.75\textwidth]{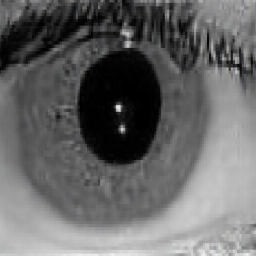}
         \caption{alpha-wolf image}
         
     \end{subfigure}
     \begin{subfigure}[b]{0.15\textwidth}
         \centering
         \includegraphics[width=\textwidth]{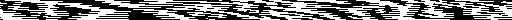}
         \caption{alpha-wolf IrisCode}
       
     \end{subfigure}
     \hfill
     \begin{subfigure}[b]{0.15\textwidth}
         \centering
         \includegraphics[width=\textwidth]{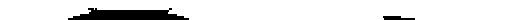}
         \caption{alpha-wolf mask}
         
     \end{subfigure} \\
     \begin{subfigure}[b]{0.12\textwidth}
         \centering
         \includegraphics[width=\textwidth]{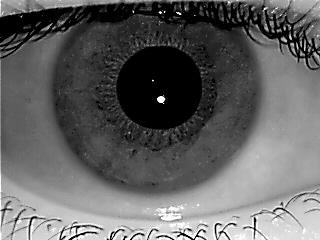}
         \caption{wolf1 image}
        
     \end{subfigure}
     \begin{subfigure}[b]{0.15\textwidth}
         \centering
         \includegraphics[width=\textwidth]{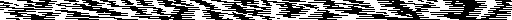}
         \caption{wolf1 IrisCode}
        
     \end{subfigure}
     \hfill
      \begin{subfigure}[b]{0.15\textwidth}
         \centering
         \includegraphics[width=\textwidth]{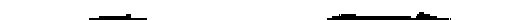}
         \caption{wolf1 mask}
       
     \end{subfigure}\\
     \begin{subfigure}[b]{0.12\textwidth}
         \centering
         \includegraphics[width=\textwidth]{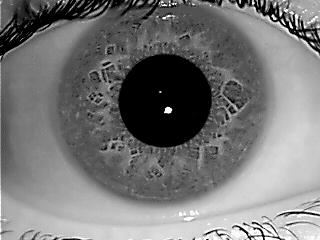}
         \caption{wolf2 image
         }
        
     \end{subfigure}
     \begin{subfigure}[b]{0.15\textwidth}
         \centering
         \includegraphics[width=\textwidth]{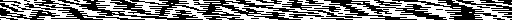}
         \caption{wolf2 IrisCode}
       
     \end{subfigure}
     \hfill
      \begin{subfigure}[b]{0.15\textwidth}
         \centering
         \includegraphics[width=\textwidth]{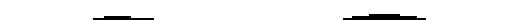}
         \caption{wolf2 mask}
        
     \end{subfigure}
        \caption{Illustration of alpha-wolf translated iris image (top row) and the respective constituent seed wolves (bottom two rows) along with their IrisCodes and masks.}
      
    \label{Fig:pix2pix}
\end{figure}

\subsection{Additional analysis}

\textbf{Is there an overlap among the seed IrisCodes across feature encoding and eye laterality?}  We observe that identical subjects (but different samples) appear as wolves for both \textit{lg} and \textit{qsw} within the same laterality (right) on the IITD dataset. For example, \textit{lg} has seed wolves \{074\textunderscore09, 150\textunderscore06\}, while \textit{qsw} has \{074\textunderscore06, 150\textunderscore09\}. XX\textunderscore YY denotes subject XX and sample YY. We note minimal overlap among the seed wolves on the CASV4-Th dataset.

\textbf{Can we assess the viability of the mixed IrisCodes at the image level?} To examine the viability of the mixed IrisCodes, we use an off-the-shelf image-to-image translation network~\cite{pix2pix} to accept a binary IrisCode as input and generate the corresponding iris image as output. We use a simple network, pix2pix~\cite{pix2piximp}, as our main objective is to inspect the viability of the mixed IrisCodes as biologically plausible ``human" iris pattern. We incorporate the Deep Image Structure and Texture Similarity (DISTS)~\cite{DISTS} index as an auxiliary term to the standard GAN expectation and $L_1$ loss terms in the formulation to account for textural and structural details preservation in the generated image. Therefore, the final generator loss function is as follows.
\begin{equation}
    \mathcal{L}_\mathcal{G} = \lambda_{GAN}*\mathcal{L}_\mathcal{GAN} + \lambda_{L1}*\mathcal{L}_{L1} + \lambda_{DISTS}*\mathcal{L}_{DISTS}
    \label{Eq:DISTS}
\end{equation}

\begin{figure}
     \centering
     \begin{subfigure}[b]{0.12\textwidth}
         \centering
         \includegraphics[width=\textwidth]{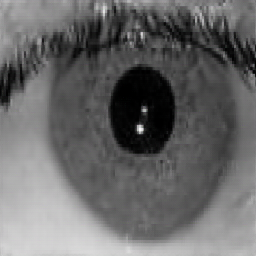}
         \caption*{}
        
     \end{subfigure}  
     \begin{subfigure}[b]{0.12\textwidth}
         \centering
         \includegraphics[width=\textwidth]{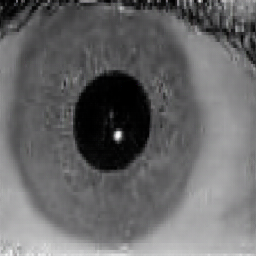}
         \caption*{Successful cases}
       
     \end{subfigure}
     \begin{subfigure}[b]{0.12\textwidth}
         \centering
         \includegraphics[width=\textwidth]{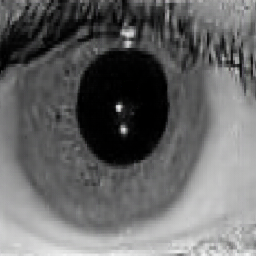}
         \caption*{}
        
     \end{subfigure} \\
\begin{subfigure}[b]{0.12\textwidth}
         \centering
         \includegraphics[width=\textwidth]{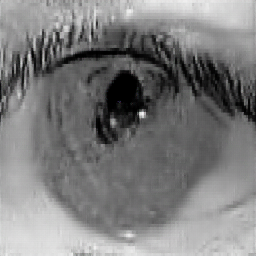}
         \caption*{}
      
     \end{subfigure}
     \begin{subfigure}[b]{0.12\textwidth}
         \centering
         \includegraphics[width=\textwidth]{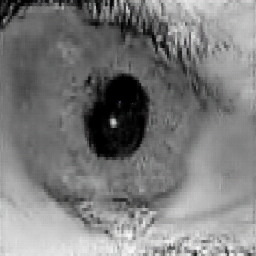}
         \caption*{Failure cases}
       
     \end{subfigure}
     \begin{subfigure}[b]{0.12\textwidth}
         \centering
         \includegraphics[width=\textwidth]{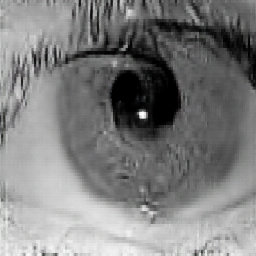}
         \caption*{}
       
     \end{subfigure}
     
    \caption{Examples of outputs from the viability check.}
    \label{Fig:moreimages}
\end{figure}

In Eqn.~\ref{Eq:DISTS}, $\lambda_{GAN}=1, \lambda_{L1}=100, \lambda_{DISTS}=0.1$. We trained the model using an ADAM optimizer with initial learning rate=$0.0002$, momentum term=$0.5$, for 200 epochs and batch size=$1$. We used the entire dataset of IITD with the original IrisCodes and the ocular images for the translation network. However, we trained separately for each eye category and feature extractor, resulting in four models (left/right $\times$ \textit{lg}/\textit{qsw}). Our test set comprises of the mixed IrisCodes (alpha-wolves) as inputs. 
See the generated images, codes and masks corresponding to the two wolves in Fig.~\ref{Fig:pix2pix}. More examples of generated iris images corresponding to alpha-wolves (mixture of wolves) are presented in Fig.~\ref{Fig:moreimages}. The image translation step filters out improbable alpha-wolves via manual inspection. Next, we recompute the user coverage with the successful alpha-wolves. We report them for XOR-mixed alpha-wolves (best-performing) in Table~\ref{Tab:postIITD}. At FMR=0.001\%, we observe an absolute decrease in the user coverage by 1.8\% on IITD left-\textit{lg}, while decreasing the number of attack attempts by $16\%$; a decrease in the user coverage by 2.7\% on IITD right-\textit{lg}, while decreasing the number of attack attempts by $39\%$; no decrease in attacks on IITD left-\textit{qsw}, while decreasing the number of attack attempts by $50\%$; and finally, a decrease in the user coverage by 13.4\% on IITD right-\textit{qsw}, while decreasing the number of attack attempts by $50\%$. 

\begin{table}[h]
\centering
\caption{User coverage after filtering alpha-wolves using viability check on IITD dataset.}
\scalebox{0.88}{
\begin{tabular}{ccccc} \hline
\multicolumn{1}{c}{\multirow{3}{*}{feature-laterality}} & \multicolumn{4}{c}{\begin{tabular}[c]{@{}c@{}}Proportion of Users (\%) Covered \\ @ False Match Rate for XOR\end{tabular}} \\ \cline{2-5}
\multicolumn{1}{c}{}  & 0.001\%   & 0.01\%  & 0.1\%  & 1\%  \\ \hline
\textit{lg}-left  & 8.9   & 16.9    & 79.0   & 89.7     \\
\textit{lg}-right & 15.2   & 17.8    & 62.5    & 72.3     \\
\textit{qsw}-left & 20.9   & 25.0    &84.4    & 93.3     \\
\textit{qsw}-right& 18.3   & 20.9    &83.0    & 87.1     \\ \hline      
\end{tabular}}
\label{Tab:postIITD}
\end{table}

\begin{table*}[t]
\centering
\caption{Cross-attacks using alpha-wolves on IITD dataset.}
\scalebox{0.75}{
\begin{tabular}{l|l|llll||llll} \hline
\multicolumn{1}{c}{\multirow{3}{*}{\begin{tabular}[c]{@{}c@{}}$\alpha_{W}$\\ $\rightarrow$\\ $\mathcal{D}_{IC}$\end{tabular}}} & \multicolumn{1}{c}{\multirow{3}{*}{\begin{tabular}[c]{@{}c@{}}Thres-\\holds\end{tabular}}}       & \multicolumn{4}{c||}{Left}                                                                                                         & \multicolumn{4}{|c}{Right}                                                                                                        \\
\multicolumn{1}{c}{}                                                                                                                                              & \multicolumn{1}{c}{}                                   & \multicolumn{4}{|c||}{\begin{tabular}[c]{@{}c@{}}Proportion of Users (\%) Covered\\ @ False Match Rate for OR/AND/XOR\end{tabular}} & \multicolumn{4}{|c}{\begin{tabular}[c]{@{}c@{}}Proportion of Users (\%) Covered\\ @ False Match Rate for OR/AND/XOR\end{tabular}} \\ \cline{3-10}
\multicolumn{1}{c}{}                                                                                                                                              & \multicolumn{1}{c}{}                                   & \multicolumn{1}{|c}{0.001}        & \multicolumn{1}{c}{0.01}        & \multicolumn{1}{c}{0.1}       & \multicolumn{1}{c||}{1}       & \multicolumn{1}{c}{0.001}        & \multicolumn{1}{c}{0.01}        & \multicolumn{1}{c}{0.1}       & \multicolumn{1}{c}{1}       \\ \hline 
\multirow{2}{*}{$lg \rightarrow qsw$}                                                                                                                & $\tau_{\alpha_{W}}$ & 0.0/0.0/10.27                    & 0.0/0.0/18.75                   & 2.23/1.79/89.73               & 2.23/1.79/92.41             & 0.0/0.0/16.96                    & 0.0/0.0/20.09                   & 1.79/2.23/87.50               & 1.79/2.23/89.73             \\
                                                                                                                                                                  & $\tau_{D_{IC}}$                    & 0.0/0.0/12.5                     & 0.0/0.0/22.70                   & 2.23/1.79/\textbf{92.86         }      & 3.57/2.23/\textbf{95.09         }    & 0.0/0.0/20.09                    & 0.0/0.0/23.21                   & 2.23/2.68/89.73               & 4.02/4.02/89.73             \\
\multirow{2}{*}{$qsw \rightarrow lg$}                                                                                                                & $\tau_{\alpha_{W}}$ & 0.0/0.0/21.43                    & 0.0/0.0/23.66                   & 1.79/2.68/89.29               & 2.23/4.02/93.30             & 0.0/0.0/\textbf{29.02}                    & 0.0/0.0/\textbf{34.82}                   & 2.23/2.68/87.95               & 2.68/3.57/89.29             \\
                                                                                                                                                                  & $\tau_{D_{IC}}$                     & 0.0/0.0/20.98                    & 0.0/0.0/22.77                   & 1.79/2.23/78.57               & 1.79/2.68/89.29             & 0.0/0.0/26.79                    & 0.0/0.0/29.02                   & 1.79/1.79/78.12               & 2.23/1.79/87.50  \\ \hline          
\end{tabular}}
\label{Tab:cross}
\end{table*}

\textbf{Can synthetic IrisCodes be used to launch dictionary attacks?}
\noindent In this experiment, we use CASIA-IrisV4-Synthetic dataset~\cite{CASIAv4} that comprises 10,000 images from 1,000 synthetic identities (no left or right eye category). Refer to~\cite{PCA, QSW} for synthesis details. We use a subset of 4,000 images (first 4 samples) from the dataset without masks.

\textbf{Intra-dataset performance:} We observe a maximum user coverage of 4.3\% with \textit{lg} and 4.0\% using \textit{qsw}, both @FMR=0.1\% by OR-ing two synthetic IrisCodes. 

\textbf{Cross-dataset performance:} On CASV4-Th left dataset, we observe @FMR=0.1\%, a maximum user coverage of 1.0\% using \textit{lg} by AND-ing two synthetic IrisCodes and 22.7\% using \textit{qsw} by OR-ing two IrisCodes. On CASV4-Th right dataset, we observe a maximum user coverage of 2.6\% using \textit{lg} by AND-ing two IrisCodes, and 47.7\% using \textit{qsw} by OR-ing two IrisCodes. We achieve a maximum user coverage of 16.8\% using \textit{lg} by XOR-ing two IrisCodes when tested on \textit{qsw}-based IrisCodes @FMR=1\%.

We also adopt a 2-state Hidden Markov Model (HMM) with a transition probability $\alpha=0.9$ as suggested in~\cite{InfoIris} to generate 10 synthetic IrisCodes. We mix them using the bitwise operators and then test them on real datasets (IITD and CASV4-Th). We observe AND-mixing achieves best user coverage of 0.4\% on the IITD dataset and 12.1\% on the CASV4-Th dataset, thus, indicating synthetic IrisCodes can be used as dictionary attacks against real IrisCodes.

         
        
        
     

\textbf{Are the attacks effective assuming limited knowledge?}
\noindent Previous experiments consider that the adversary has full knowledge of the feature encoding scheme and their corresponding decision thresholds. In this experiment, we use $lg$-based alpha-wolves, $\alpha_W$ (from 2 wolves) to launch dictionary attacks against $qsw$-based IrisCodes, $\mathcal{D}_{IC}$, and vice-versa. We compute the user coverage considering multiple thresholds, $\tau_{\alpha_{W}}$ (threshold corresponding to the feature of the alpha-wolves) and $\tau_{D_{IC}}$ (threshold corresponding to the feature of the target IrisCode). This experiment assumes limited knowledge on part of the adversary about the encoding employed by the target system. Results in Table~\ref{Tab:cross} indicate that even with partial knowledge, template level attacks achieve an extremely high coverage of 29.02\% @FMR=0.001\% when XOR-mixed $qsw$-based wolves are used against $lg$-based IrisCodes on the IITD right dataset. 

\textbf{Can we learn the \textit{best} possible way to combine IrisCodes?}
We have a fixed set of logical operators as the mixing function but that does not guarantee optimal mixing. Therefore, we employ an existing image fusion technique, namely IFCNN~\cite{IFCNN} to perform mixing. IFCNN uses a 4-layer CNN trained on over 100K RGB and depth-images for fusing multi-modal, multi-spectral and multi-exposure images using MSE and perceptual losses extracted from a pre-trained ResNet 101 model. We selected this network as it allows fusion of variable number of inputs. We supply seed wolves as templates in one setup and as iris images in another setup to perform mixing at both template and image level. The best user coverage from the fused wolf IrisCodes is 21.9\%, and from the fused wolf iris images is 17.4\% @FMR=1\% after fusing 2 codes/images. 


\textbf{Summary:}
We study the feasibility of dictionary attacks on iris recognition systems for the first time. Although the practicality of this attack is currently restricted at the template (IrisCode) level, we observe vulnerabilities on high quality iris datasets and we suspect that the risk might be further compounded in the presence of non-ideal imaging (low resolution, inadequate illumination, etc.). Our findings surprisingly indicate that mixing IrisCodes using simple bitwise operators can be highly effective as dictionary attacks against a large number of unseen identities. We observe that, in particular, XOR-operator increases the user coverage, \textit{e.g.}, from 1.34\% on IITD left-$lg$ wolves to 20.5\% @FMR=0.01\% with XOR-mixed alpha-wolves (only wolf samples) and 51.6\% @FMR=0.1\% with XOR-mixed alpha-mammals (with or w/o wolves). Even synthetic IrisCodes can be used as alpha-wolves to launch dictionary attacks on real datasets with 47.7\% coverage @FMR=0.1\%. We further validate the viability of alpha-mixtures at image level via a conditional generative network. 

\section{Conclusion} 
In this work, we explore dictionary attacks at the template level (IrisCodes) on iris recognition systems that use log Gabor (\textit{lg}) and spatial Gabor (\textit{qsw}) features. We show that mixing IrisCodes using AND, OR and XOR operators results in the so-called Master IrisCodes that can fortuitously match with a large number of other identities. The IrisCodes are strategically selected: they can be either wolves, resulting in alpha-wolves, or selected via search optimization, resulting in alpha-mammals. We empirically analyze the efficacy of these attacks on three datasets, \textit{viz.}, IITD, CASIA-IrisV4-Thousand and Synthetic, and achieve a user coverage of upto 71 identities @FMR=0.001\% using real alpha-wolves, upto 133 identities using alpha-mammals at FMR=0.1\% and upto 477 identities @FMR=0.1\% using synthetic alpha-wolves. Our method is effective on cross-attacks across different IrisCode encoding schemes. Future work will extend to image-level dictionary attacks. 


\clearpage
\balance
{\small
\bibliographystyle{ieee_fullname}
\bibliography{egbib}
}

\end{document}